\documentclass[sigconf,nonacm]{acmart}
\makeatletter
\def\@ACM@checkaffil{
    \if@ACM@instpresent\else
    \ClassWarningNoLine{\@classname}{No institution present for an affiliation}%
    \fi
    \if@ACM@citypresent\else
    \ClassWarningNoLine{\@classname}{No city present for an affiliation}%
    \fi
    \if@ACM@countrypresent\else
        \ClassWarningNoLine{\@classname}{No country present for an affiliation}%
    \fi
}
\makeatother
\AtBeginDocument{%
  \providecommand\BibTeX{{%
    \normalfont B\kern-0.5em{\scshape i\kern-0.25em b}\kern-0.8em\TeX}}}

\setcopyright{acmcopyright}
\copyrightyear{2024}
\acmYear{2023}
\acmDOI{10.1145/1122445.1122456}

\acmConference[KDD '24]{KDD '24: 30th ACM SIGKDD Conference on Knowledge Discovery & Data Mining}{Aug 25 - 29, 2024}{Barcelona, Spain}
\acmBooktitle{KDD '24: 29th ACM SIGKDD Conference on Knowledge Discovery & Data Mining,
  Aug 25 - 29, 2024, Barcelona, Spain} 

\acmPrice{15.00}
\acmISBN{978-1-4503-XXXX-X/18/06}

\usepackage{multirow}
\usepackage{amsmath}
\usepackage{amsfonts}
\usepackage{color}
\usepackage{graphicx}
\usepackage{enumitem}
\usepackage{listings}
\usepackage{bbding}
\usepackage{subcaption}
\usepackage{diagbox}
\usepackage{paralist}
\usepackage[normalem]{ulem}
\useunder{\uline}{\ul}{}

\usepackage[toc,page]{appendix}

\usepackage{tikz}
\usepackage[edges]{forest}
\definecolor{hidden-draw}{RGB}{0,0,0}
\definecolor{hidden-pink}{rgb}{0.98, 0.94, 0.75}
\definecolor{level0}{rgb}{0.67, 0.88, 0.69}
\definecolor{level1}{rgb}{0.98, 0.92, 0.84}
\definecolor{level2}{rgb}{0.8, 0.8, 1.0}
\definecolor{level3}{rgb}{1.0, 0.71, 0.76}
\definecolor{level4}{rgb}{0.49, 0.99, 0.0}

\definecolor{lawngreen}{rgb}{0.49, 0.99, 0.0}
\definecolor{pink}{rgb}{1, 0, 0.5}
\definecolor{airforce}{rgb}{0.36, 0.54, 0.66}

\hypersetup{
    linkbordercolor=airforce,
    urlbordercolor=pink,
    citebordercolor=green,
}

\begin{document}

\title{Large Language Models for Education: A Survey and Outlook}



		





\author{Shen Wang\textsuperscript{\rm 1}$^{\ast}$, Tianlong Xu\textsuperscript{\rm 1}$^{\ast}$, Hang Li\textsuperscript{\rm 2}$^{\ast}$, Chaoli Zhang\textsuperscript{\rm 3}, Joleen Liang\textsuperscript{\rm 4},\\ Jiliang Tang\textsuperscript{\rm 2}, Philip S. Yu\textsuperscript{\rm 5}, Qingsong Wen\textsuperscript{\rm 1}$^{\dagger}$}
\thanks{$^{\ast}$Equal contribution. $^{\dagger}$Corresponding author.}
\affiliation{%
  \institution{\textsuperscript{\rm 1}Squirrel AI, USA \hspace{0.2em}
  \textsuperscript{\rm 2}Michigan State University, USA \hspace{0.2em}
  \textsuperscript{\rm 3}Zhejiang Normal University, China \hspace{0.2em}\\
  \textsuperscript{\rm 4}Squirrel AI, China \hspace{0.2em}
   \textsuperscript{\rm 5}University of Illinois Chicago, USA \\
  }
}

\begin{abstract}
The advent of large language models (LLMs) has brought in a new era of possibilities in the realm of education. This survey paper summarizes the various technologies of LLMs in educational settings from multifaceted perspectives, encompassing student and teacher assistance, adaptive learning, and commercial tools. We systematically review the technological advancements in each perspective, organize related datasets and benchmarks, and identify the risks and challenges associated with the deployment of LLMs in education. Furthermore, we outline future research opportunities, highlighting the potential promising directions. Our survey aims to provide a comprehensive technological picture for educators, researchers, and policymakers to harness the power of LLMs to revolutionize educational practices and foster a more effective personalized learning environment.
\end{abstract}

\maketitle

\section{Introduction}

During the past decades, artificial intelligence (AI) for education has received a great deal of interest and has been applied in various educational scenarios~\cite{chen2020artificial,maghsudi2021personalized,chiu2023systematic,denny2024computing,li2024bringing, latif2023artificial}. Specifically, educational data mining methods have been widely adopted in different aspects such as cognitive diagnosis, knowledge tracing, content recommendations, as well as learning analysis~\cite{romero2007educational,romero2010educational,romero2013data,koedinger2015data,romero2020educational,batool2023educational,xiong2024review}. 

As large language models (LLMs) have become a powerful paradigm in different areas~\cite{fan2023recommender,zeng2023large,jin2024position,chen2023exploring}, they also achieved state-of-the-art performances in multiple educational scenarios~\cite{li2023adapting,kasneci2023chatgpt,yan2024practical}.
Existing work has found that LLMs can achieve student-level performance on standardized tests \cite{openai2023gpt} in a variety of mathematics subjects (e.g., physics, computer science) on both multiple-choice and free-response problems. In addition, empirical studies have shown that LLMs can serve as a writing or reading assistant for education \cite{malinka2023educational,susnjak2022chatgpt}. A recent study \cite{susnjak2022chatgpt} reveals that ChatGPT is capable of generating logically consistent answers across disciplines, balancing both depth and breadth. Another quantitative analysis \cite{malinka2023educational} shows that students using ChatGPT (by keeping or refining the results from LLMs as their own answers) perform better than average students in some courses from the field of computer security. Recently, several perspective papers \cite{tan2023towards, kamalov2023new} also explore various application scenarios of LLMs in classroom teaching, such as teacher-student collaboration, personalized learning, and assessment automation. However, the application of LLMs in education may lead to a series of practical issues, e.g., plagiarism, potential bias in AI-generated content, overreliance on LLMs, and inequitable access for non-English speaking individuals~\cite{kasneci2023chatgpt}.

To provide researchers with a broad overview of the domain, numerous exploratory and survey papers have been proposed. For example, \citet{qadir2023engineering, rahman2023chatgpt} and \citet{rahman2023chatgpt} conclude the applications of ChatGPT to engineering education by analyzing the responses of ChatGPT to the related pedagogical questions. \citet{jeon2023large} and \citet{mogavi2023exploring} collect the opinions from different ChatGPT user groups, e.g., educator, learner, researcher, through in-person interviews, online post replies and user logs, and conclude the practical applications of LLMs in education scenarios. \citet{baidoo2023education} and \citet{zhang2023systematic} focus on the literature review over the published papers and summarize the progress of the area with structured tables. Although the above works have covered a wide range of existing applications of LLMs in education scenarios and provided their long-term visions for future studies, we argue that none of the literature has systematically summarized LLMs for education from a technological perspective.
To bridge this gap, this survey aims to provide a comprehensive technological review of LLMs for education, which provides a novel technology-centric taxonomy and summary of existing publicly available datasets and benchmarks. Furthermore, we summarize current challenges as well as further research opportunities, in order to foster innovations and understanding in the dynamic and ever-evolving landscape of LLMs for education. In summary, our contributions lie in the following three main parts:

\noindent    \textbf{1, Comprehensive and up-to-date survey.} We offer a comprehensive and up-to-date survey on LLMs for a wide spectrum of education, containing academic research, commercial tools, and related datasets and benchmarks.
    
\noindent    \textbf{2, New technology-centric taxonomy.} We provide a new taxonomy that offers a thorough analysis from a technological perspective on LLMs for education, encompassing student and teacher assistance, adaptive learning, and commercial tools.

\noindent    \textbf{3, Current challenges and future research directions} We discuss current risks and challenges, as well as highlight future research opportunities and directions, urging researchers to dive deeper into this exciting area.

 \vspace{-5pt}
\section{LLM in Education Applications}

\tikzstyle{my-box}=[
    rectangle,
    draw=hidden-draw,
    rounded corners,
    text opacity=1,
    minimum height=1.5em,
    minimum width=5em,
    inner sep=2pt,
    align=center,
    fill opacity=.5,
    line width=0.8pt,
]
\tikzstyle{leaf}=[my-box, minimum height=1.5em,
    fill=hidden-pink!80, text=black, align=left, font=\normalsize,
    inner xsep=2pt,
    inner ysep=4pt,
    line width=0.8pt,
]
\begin{figure*}[!t]
    \centering
    \resizebox{0.88\textwidth}{!}{
        \centering
\includegraphics[width =1\linewidth]{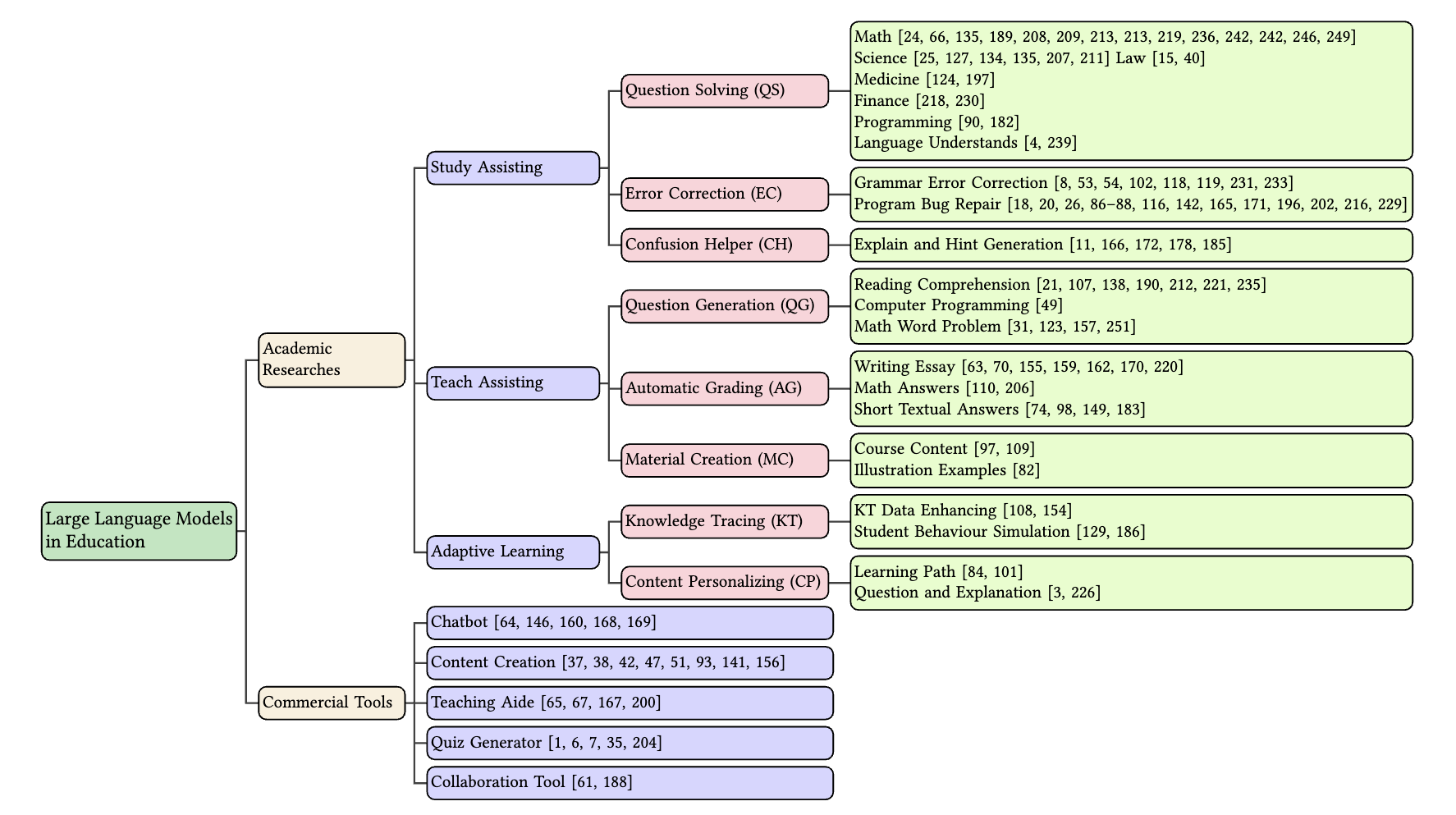}
    }\vspace{-3mm}
\caption{A taxonomy of LLMs for education applications with representative works.} \vspace{-4mm}
\label{fig:taxonomy}
\end{figure*}
\subsection{Overview}

The education application can be categorized based on its users' role in education and its usage scenario in education. In this paper, we summarize the appearance of LLMs in different applications and discuss the benefits brought by LLMs compared to the original methods. We present our primary summary of education applications with LLMs using a taxonomy illustrated in Figure~\ref{fig:taxonomy}.


\vspace{-5pt}
\subsection{Study Assisting}

Providing students with timely learning support has been widely recognized as a crucial factor in improving student engagement and learning efficiency during their independent studies \cite{dewhurst2000independent}. Due to the limitation of prior algorithms in generating fixed-form responses, many of the existing study-assisting approaches face poor generalization challenges while being implemented in real-world scenarios \cite{konig2023critical}. Fortunately, the appearance of LLMs brings revolutionary changes to this field. Using finetuned LLMs \cite{ouyang2022training} to generate human-like responses, recent studies in LLM-based educational support have demonstrated promising results. These studies provide real-time assistance to students by helping them solve challenging questions, correcting errors, and offering explanations or hints for areas of confusion.




\subsubsection{Question Solving (QS)}

Contributing to the large-scale parameter size of LLMs and the enormous sized and diverse web corpus used during the pre-training phase, LLMs have been proven to be a powerful question zero-shot solver to questions spread from a wide spread of subjects, including math \cite{yuan2023well,wu2023empirical}, law \cite{bommarito2022gpt,cui2023chatlaw}, medicine \cite{thirunavukarasu2023large,lievin2023can}, finance \cite{wu2023bloomberggpt,yang2023fingpt}, programming \cite{kazemitabaar2023novices,savelka2023thrilled}, language understands\cite{zhang2024m3exam,achiam2023gpt}. In addition, to further improve LLM's problem-solving performance while facing complicated questions, a variety of studies have been actively proposed. For example, \citet{wei2022chain} propose the Chain-of-Thought (CoT) prompting method, which guides LLMs to solve a challenging problem by decomposing it into simpler sequential steps. Other works \cite{sun2023offline,wang2023large} exploit the strong in-context learning ability of LLMs and propose advanced few-shot demonstration-selection algorithms to improve LLM's problem-solving performance to general questions. \citet{chen2022program} and \citet{gao2023pal} leverage external programming tools to avoid calculation errors introduced during the textual problem solving process of raw LLMs. \citet{wu2023autogen} regard chat-optimized LLMs as powerful agents and design a multi-agent conversation to solve those complicated questions through a collaborative process. \citet{cobbe2021training} and \citet{zhou2023solving} propose the external verifier module to rectify intermediate errors during generation, which improves LLM problem solving performance on challenging math questions. Overall, with the proposition of all these novel designs, the use of LLMs for question solving has achieved impressive progress. Furthermore, students can find high-quality answers to their blocking questions in a timely manner.







\subsubsection{Error Correction (EC)}


Error correction focuses on providing instant feedback to students' errors that they make during the learning process. This action is helpful for students in the early stages of learning. \citet{zhang2023does} explore using four prompt strategies: zero-shot, zero-shot-CoT, few-shot, and few-shot-CoT to correct common grammar errors in Chinese and English text. From their experiment, they find that LLMs have tremendous potential in the correction task, and some simple spelling errors have been perfectly solved by the current LLMs. GrammarGPT \cite{fan2023grammargpt} leverages LLM for addressing native Chinese grammatical errors. By fine-tuning open-source LLMs with hybrid annotated dataset, which involves both human annotation and ChatGPT generation, the proposed framework performs effectively in native Chinese grammatical error correction. \citet{zhang2022repairing} propose to use a large language model trained in code, such as Codex, to build an APR system – MMAPR – for introductory Python programming assignments. With MMAPR evaluated on real student programs and comparing it to the prior state-of-the-art Python syntax repair engine, the author found that MMAPR can fix more programs and produce smaller patches on average. \citet{do2023using} develop a few-shot example generation pipeline, which involves code summarize generation and code modification to create few-shot examples. With the generated few-shot examples, the performance and stability of LLMs on bug fixing performance on student programs receive a great boost. 

\subsubsection{Confusion Helper (CH)}


Unlike QS and AC, studies in the confusion helper direction avoid providing correct problem solutions directly. Instead, these works aim to use LLMs to generate pedagogical guidance or hints that help students solve problems themselves. \citet{shridharautomatic} propose various guided question generation schemes based on input conditioning and reinforcement learning, and explore the ability of LLMs to generate sequential questions to guide the solution of math word problems. \citet{prihar2023comparing} explore using LLMs to generate explanations for math problems in two ways: summarizing question-related tutoring chat logs and learning a few shots from existing explanation text. Based on their experiment, they find the synthetic explanations cannot outperform teacher-written explanations, as some terms may not be known by students, and the advice is sometimes too general. The research by \citet{pardos2023learning} evaluates the learning gain differences between ChatGPT and human-tutor-generated algebra hints. By observing the changes in participants' pre-test and post-test scores between the controlled groups, the author draws a similar conclusion that hints generated by LLMs are less effective in guiding students to find worked solutions. \citet{balse2023evaluating} evaluate the validity of using LLMs to generate explaining text to logical errors in students' computer programming homework. By ranking the synthetic explanations between the ones written by course TAs, the author finds that synthetic explanations are competitive to human-generated results but have shortages in correctness and information missing problems. \citet{rooein2023know} experiment with generating adaptive explanations for different groups of students. By introducing controlling conditions, such as age group, education level, and detail level, to the instructional prompt, the proposed method adapts the generated explanations to students with diverse learning profiles.

\begin{figure*}[!btph]
    \centering
    \includegraphics[width=0.95\textwidth]{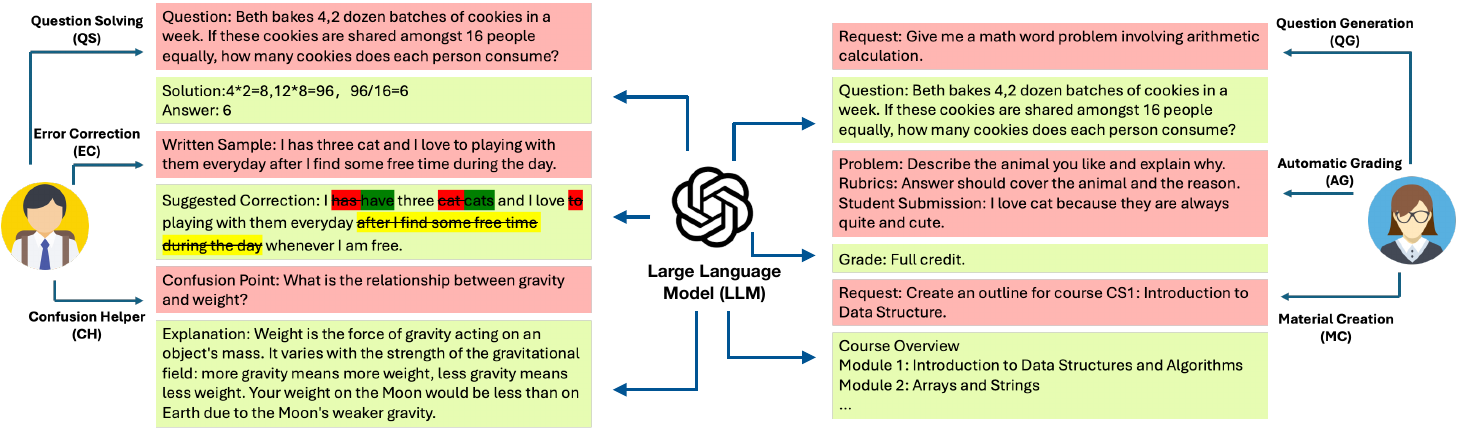}
    \caption{LLMs in student and teacher assisting.}
    \label{fig:student_teacher}
\end{figure*}

\vspace{-5pt}
\subsection{Teach Assisting}

Contributing to LLM's unprecedented logical reasoning and problem-solving capability, developing LLM-based teach-assisting models has become another popular topic in education research recently. With the help of these assisting algorithms, instructors are able to get rid of prior burdensome routine workloads and focus their attention on tasks like in-class instructions, which cannot be replaced by existing machine learning models.





\subsubsection{Question Generation (QG)}

Due to its highly frequent usage in pedagogical practice, Question Generation (QG) has become one of the most popular research topics in LLMs' application for education. \citet{xiao2023evaluating} leverage LLMs to generate reading comprehension questions by first fine-tuning it with supplemental reading materials and textbook exercise passages, then by employing a plug-and-play controllable text generation approach, fine-tuned LLMs are guided in generating more coherent passages based on specified topic keywords. \citet{doughty2024comparative} analyze the ability of LLM (GPT-4) to produce multiple-choice questions (MCQs) aligned with specific learning objectives (LO) of Python programming classes in higher education. By integrating several generation control modules with the prompt assembly process, the proposed framework is capable of producing MCQs with clear language, a single correct choice, and high-quality distractors. \citet{lee2023few} focused on aligning prompting questions and reading comprehension taxonomy using a 2D matrix-structured framework. Using aligned prompts, LLM questions can cover a broad range of question types and difficulty levels in a balanced manner. \citet{zhou2023learning} work on generating diverse math word problems with implicit diversity controls toward the equation of question and achieve the goal of generating high-quality diverse questions.

\subsubsection{Automatic Grading (AG)}

Research on automatic assignment graders has been proposed much earlier to the recent emergence of LLMs. However, due to the limitations in the learning capability of prior models, most existing auto-grading algorithms \cite{liu2019automatic} focus on exploring semantic comparisons between golden solutions and student responses, which dismisses the logical considerations behind manual scoring processes. Apart from that, as the quality of the provided solution heavily influences the result, the applications of previous works are restricted to some well-annotated problems. Fortunately, with the appearance of LLMs, the above challenges have become easy to solve. Studies by \citet{yancey2023rating} and \citet{pinto2023large} first explore the usage of LLMs for automatic scoring in open-ended questions and writing essays using prompt tuning algorithms. By including comprehensive contexts, clear rubrics, and high-quality examples, LLMs demonstrate its satisfactory performance on both grading tasks. \citet{xiao2024automation} further integrates CoT within the grading process. This approach instructs LLMs to first analyze and explain the provided materials before making final score determinations. With such modifications, the LLMs will not only generate the score results, but also provide detailed comments to the students' responses, which helps students to learn how to improve for the next time. \citet{li2024automated} extend the grading object from the textual responses of the students to those with handwritten responses. With the use of the advanced multimodal LLM framework, for example, CLIP and BLIP, the work demonstrates that the incorporation of the student's text and image, as well as the text and image of the question, improves the model's grading performance. \citet{funayama2023reducing} propose a pre-finetuning method for the cross-prompt to learn the shared relationship between different rubrics and annotated examples, then by further tuning the pre-finetuned LLMs on the target scoring task, the model can achieve comparable performance under the limitation of labeled samples.

\subsubsection{Material Creation (MC)}

Despite the above tasks, pioneering researchers also find the great potential of LLMs to help teachers create high-quality educational materials. For example, \citet{leiker2023prototyping} presents an investigation into the use of LLMs in asynchronous course creation, particularly within the context of adult learning, training, and upskilling. To ensure the accuracy and clarity of the generated content, the author integrates LLMs with a robust human-in-the-loop process. \citet{koraishi2023teaching} leverage GPT-4 with a zero-shot prompt strategy to optimize the materials of the English as a Foreign Language (EFL) course. In their exploration, the authors examine how ChatGPT can be used in material development, streamlining the process of creating engaging and contextually relevant resources tailored to the needs of individual learners, as well as other more general uses. \citet{jury2024evaluating} present a novel tool, 'WorkedGen', which uses LLMs to generate interactive worked examples. Through the use of strategies such as prompt chaining and one-shot learning to optimize the output, the generated work examples receive positive feedback from students.

\vspace{-5pt}
\subsection{Adaptive Learning}



Based on the specific problems solved by the proposed methods, existing work on adaptive learning can be classified into two categories: knowledge tracing \cite{abdelrahman2023knowledge} and content personalization \cite{naumov2019deep}. To be specific, knowledge tracing targets estimating the students' knowledge mastery status based on the correctness of students' responses to questions during their study processes. Content personalizing focuses on providing customized learning content to students based on personalized factors such as learning status, preferences, and goals. During the past few decades, a variety of machine learning algorithms, including traditional statistical methods \cite{kuvcak2018machine} and advanced deep learning models \cite{lin2023comprehensive}, have been explored by different studies, and some promising results have been achieved for both problems \cite{liu2017investigating}. With the recent surge of powerful LLMs in various applications, novel opportunities are also emerging for research in these directions.

\subsubsection{Knowledge Tracing (KT)}

The current usage of LLMs in knowledge tracing focuses on generating auxiliary information to both the question text and the student records data. In recent work by \citet{ni2023enhancing}, the author uses LLM to extract the knowledge keyword for each text of questions in the student-question response graph. Contributing to LLM’s strong generous capabilities to deal with unseen text, the proposed framework proves especially advantageous in addressing cold start scenarios characterized by limited student question practice data. In addition to that, \citet{lee2023difficulty} proposes a framework, DCL4KT+LLM, which predicts the difficulties of questions based on the text of the question stem and the concepts of knowledge associated with LLM. Using the predicted question difficulties, DCL4KT+LLM overcomes the missing difficulty information problem of existing knowledge tracing algorithms when faced with unseen questions or concepts. Finally, \citet{sonkar2023deduction} explores the capabilities of LLM in logical reasoning with distorted facts. By utilizing the prompts designed by the study, the LLMs demonstrate the possibility of simulating students' incorrect responses when given the appropriate knowledge profiles of the students.

\subsubsection{Content Personalizing (CP)}

As most advanced LLMs are generative models, the use of LLMs to create personalized learning content has been explored in many recent education researches. For example, \citet{kuo2023leveraging} attempts to generate a dynamic learning path for students based on their most recent knowledge mastery diagnosis result. \citet{kabir2023llm} incorporate the knowledge concept structures during the generation. Specifically, if the student masters a topic for a given Learning Object (LO), a question from the next LO will be automatically generated. \citet{yadav2023contextualizing} explore the potential of LLMs in creating contextualizing algebraic questions based on student interests. By conducting iterative prompt engineering on the few-shot learning approach, the system aptly accommodates novel interests such as TikTok and NBA to the generated question stem text, which helps to improve student engagement and outcomes during the study. Despite generating content, other studies \cite{abu2024knowledge} also try to leverage chat-based LLMs to generate explanations of learning recommendations. By utilizing Knowledge Graphs (KGs) as a source of contextual information, the approach demonstrates its capability to generate convincing answers for a learner who has inquiries about the learning path recommended by ITS systems.

\vspace{-5pt}
\subsection{Education Toolkit}

Besides leveraging LLMs to empower well-formulated education applications in academia, several LLM-powered commercial education tools have been developed in the industry. In particular, they can be categorized into five categories, including Chatbot, Content Creation, Teaching Aide, Quiz Generator and Collaboration Tool.

\subsubsection{Chatbot}
Using a Large Language Model (LLM) chatbot as an educational tool offers a range of advantages and opportunities. LLM chatbots can adapt their responses to the individual needs of learners, providing personalized feedback and support. This customization can accommodate different learning styles, speeds, and preferences. They offer 24/7 availability, making learning accessible anytime, anywhere. This can be particularly beneficial for learners in different time zones or with varying schedules. The interactive nature of chatbots can make learning more engaging and fun. They can simulate conversations, create interactive learning scenarios, and provide instant feedback, which can be more effective than passive learning methods. Chatbots can handle thousands of queries simultaneously, making them a scalable solution for educational institutions to support a large number of learners without a corresponding increase in teaching staff. They can automate repetitive teaching tasks, such as grading quizzes or providing basic feedback, allowing educators to focus on more complex and creative teaching responsibilities. There are some representative chatbots, such as ChatGPT \cite{OpenAIChat}, Bing Chat \cite{Microsoft2023EnhanceTeaching},  Google Bard \cite{GoogleBard}, Perplexity \cite{PerplexityAI}, Pi \citet{PiAI}.

\subsubsection{Content Creation}
Curipod \cite{Curipod} takes user input topics and generates an interactive slide deck, including polls, word clouds, open-ended questions, and drawing tools. Diffit \cite{Diffit} provides a platform on which the user can find leveled resources for virtually any topic. It enables teachers to adapt existing materials to suit any reader, create customized resources on any subject, and then edit and share these materials with students. MagicSchool \cite{MagicSchoolAI} is an LLM-powered educational platform that aims to help teachers save time by automating tasks such as lesson planning, grading, and creating educational content. It provides access to more than 40 AI tools, which can be searched by keywords and organized into categories for planning, student support, productivity, and community tools. Education Copilot \cite{Education_Copilot} offers LLM-generated templates for a variety of educational needs, including lesson plans, writing prompts, handouts, student reports, project outlines, and much more, streamlining the preparation process for educators. Nolej \cite{Nolej} specializes in creating a wide range of interactive educational content, including comprehensive courses, interactive videos, assessments, and plug-and-play content, to enhance the learning experience. Eduaide.ai \cite{Eduaide_ai} is an LLM-powered teaching assistant created to support teachers in lesson planning, instructional design, and the creation of educational content. It features a resource generator, teaching assistant, feedback bot, and AI chat, providing comprehensive assistance for educators. Khanmigo \cite{Khanmigo}, developed by Khan Academy, is an LLM-powered learning tool that serves as a virtual tutor and debate partner. It can also assist teachers in generating lesson plans and handling various administrative tasks, enhancing both learning and teaching experiences. Copy.ai \cite {Copy_ai}is an LLM-powered writing tool that uses machine learning to produce a wide range of content types, such as blog headlines, emails, social media posts, and web copy. 


\subsubsection{Teaching Aide}

gotFeedback \cite{gotFeedbackbygotLearning} is developed to assist teachers in providing more personalized and timely feedback to their students, seamlessly integrating into the gotLearning platform. It is based on research emphasizing that effective feedback should be goal-referenced, tangible and transparent, actionable, user-friendly, timely, ongoing, and consistent, ensuring that it meets students' needs effectively. 
Grammarly \cite{Grammarly} serves as an online writing assistant, employing LLM to help students write bold, clear, and error-free writing. Grammarly's AI meticulously checks grammar, spelling, style, tone, and more, ensuring your writing is polished and professional.
Goblin Tools \cite{Goblin_Tools} offers a suite of simple single-task tools specifically designed to assist neurodivergent individuals with tasks that may be overwhelming or challenging. This collection includes Magic ToDo, Formalizer, Judge, Estimator, and Compiler, each tool catering to different needs and simplifying daily tasks to enhance productivity and ease.
ChatPDF \cite{Chat_PDF} is an LLM-powered tool designed to let users interact with PDF documents through a conversational interface. This innovative approach enables easier navigation and interaction with PDF content, making it more accessible and user-friendly.

\subsubsection{Quiz Generator}

QuestionWell \cite{QuestionWell} is an LLM-based tool that generates an unlimited supply of questions, allowing teachers to focus on what is most important. By entering reading material, AI can create essential questions, learning objectives, and aligned multiple choice questions, streamlining the process of preparing educational content and assessments. 
Formative \cite{Formative_AI},  a platform for assignments and quizzes that accommodated a wide range of question types, has now enhanced its capabilities by integrating ChatGPT. This addition enables the generation of new standard-aligned questions, hints for learners, and feedback for students, leveraging the power of LLM to enrich the educational experience and support customized learning paths.
Quizizz AI \cite{Quizizz_AI} is an LLM-powered feature that specializes in generating multiple-choice questions, which has the ability to automatically decide the appropriate number of questions to generate based on the content supplied. Furthermore, Quizizz AI can modify existing quizzes through its Enhance feature, allowing for the customization of activities to meet the specific needs of students.
Conker \cite{Conker} is a tool that enables the creation of multiple-choice, read-and-respond, and fill-in-the-blank quizzes tailored to students of various levels on specific topics. It also supports the usage of user input text, from which it can generate quizzes, making it a versatile resource for educators aiming to assess and reinforce student learning efficiently.
Twee \cite{Twee} is an LLM-powered tool designed to streamline lesson planning for English teachers, generating educational content, including questions, dialogues, stories, letters, articles, multiple choice questions, and true/false statements. This comprehensive support helps teachers enrich their lesson plans and engage students with a wide range of learning materials.

\subsubsection{Collaboration Tool}
summarize.tech \cite{summarize_tech} is a ChatGPT-powered tool that can summarize any long YouTube video, such as a lecture, a live event, or a government meeting.
Parlay Genie \cite{Parlay_Genie} serves as a discussion prompt generator that creates higher-order thinking questions for classes based on a specific topic, a YouTube video, or an article. It uses the capabilities of ChatGPT to generate engaging and thought-provoking prompts, facilitating deep discussions and critical thinking among students.

  \vspace{-5pt}
\section{Dataset and Benchmark}

LLMs revolutionized the field of natural language processing (NLP) by enabling a wide range of text-rich downstream tasks, which leverage the extensive knowledge and linguistic understanding embedded within LLMs to perform specific functions requiring comprehension, generation, or text transformation. Therefore, many datasets and benchmarks are constructed for text-rich educational downstream tasks. The majority of datasets and benchmarks lie in the tasks of question-solving (QS), error correction (EC), question generation (QG), and automatic grading (AG), which cover use cases that benefit different users, subjects, levels, and languages. Some of these datasets mainly benefit the student, while others help the teacher.

Datasets and benchmarks for educational applications vary widely in scope and purpose, targeting different aspects of the educational process, such as student performance data \cite{ray2003pisa}, text and resource databases \cite{brooke2015gutentag}, online learning data \cite{ruiperez2022large}, language learning database \cite{tiedemann2020tatoeba}, education game data \cite{liu2020using}, demographic and socioeconomic data \cite{cooper2020demographic}, learning management system (LMS) data \cite{conijn2016predicting}, special education and needs data \cite{morningstar2017examining}. Specifically, the question-solving ones \cite{cobbe2021training,hendrycks2020measuring,huang2016well,wang2017deep,zhao2020ape210k,amini2019mathqa,miao2021diverse,lu2021iconqa,kim2018textbook,lu2021inter,chen2023theoremqa}, take a significant amount as a prevalent task for both Education and NLP fields. 
In particular, many datasets \cite{cobbe2021training,hendrycks2020measuring,huang2016well,wang2017deep,zhao2020ape210k,amini2019mathqa,miao2021diverse,lu2021iconqa} are constructed for math question-solving, aiming to provide an abstract expression from a narrative description. Some datasets also take the images \cite{miao2021diverse,kim2018textbook,lu2021inter,kembhavi2016diagram} and tables \cite{lu2021iconqa} into consideration. On the other hand, another bunch of datasets and benchmarks \cite {kim2018textbook,kembhavi2016diagram,chen2023theoremqa}, are constructed for science textbook question-solving, which requires a comprehensive understanding of the textbook and provide the answer corresponding to the key information in the question.
There are also large amounts of datasets and benchmarks constructed for error correction. They are used for foreign language training \cite{rothe2021simple,ng2014conll,bryant2019bea,tseng2015introduction,zhao2022overview,xu2022fcgec,du2023flacgec,naplava2022czech,rozovskaya2019grammar,grundkiewicz2014wiked,davidson2020developing,syvokon2021ua,cotet2020neural} and computer science programming language training \cite{just2014defects4j,le2015manybugs,lin2017quixbugs,tufano2019empirical,li2022automating,guo2024exploring}. The foreign language training datasets and benchmarks contains grammatical errors and spelling errors that are needed to be identified and corrected. The programming training datasets and benchmarks include several code bugs that require sufficient coding understanding for proper correction. 
On the other hand, there are several datasets and benchmarks for teacher-assisting tasks. \cite{welbl2017crowdsourcing,lai2017race,xu2022fantastic,chen2018learningq,gong2022khanq,hadifar2023eduqg,liang2018distractor,bitew2022learning} are constructed for question construction task that aims to evaluate the ability of the LLM for generating the educational questions from a given context. \cite{yang2023fingpt,tigina2023analyzing,blanchard2013toefl11,stab2014annotating} are built for automatic grading the student assignments.
We summarize the commonly used publicly available datasets and benchmarks for evaluating LLMs on education applications in Table \ref{tab:dataset} of the Appendix\footnote{While there exist various educational applications, including those that assist with confusion helper, material creation, knowledge tracking, and content personalizing, we did not discuss them in this section due to the absence of publicly accessible datasets.}.


  \vspace{-5pt}
\section{Risks and Potential Challenges}
This section discusses risks and challenges along with the rise of generative AI and LLMs and summarizes some early proposals for the implementation of guardrails and responsible AI. Given the importance of education as a critically important domain, more caution should be used when implementing the implications of LLMs. A well-established framework on responsible AI \cite{Microsoft2024ResponsibleAI} has outlined six foundational elements: fairness, inclusiveness, reliability \& safety, privacy \& security, transparency, and accountability. Besides these, for education domain, the concern of overreliance is also a major concern as over-depending on LLMs will harm some key capabilities of students such as critical thinking, academic writing, and even creativity.


\subsection{Fairness and Inclusiveness}\vspace{-1mm}
Limited by LLM training data, where representations of specific groups of individuals and social stereotypes might be dominant, bias could develop \cite{zhuo2023red}. Li et al. \cite{li2024survey} summarized that for the education domain, critical LLM fairness discussions are based on demographic bias and counterfactual concerns.
Fenu et al. \cite{fenu2022experts} has introduced some bias LLMs that fail to generate as much useful content for some groups of people not represented in the data. Also of concern is the fact that people in some demographic groups may not have equal access to educational models of comparable quality levels. Weidinger et al. \cite{weidinger2021ethical} shows the lack of LLM ability in generating content for groups whose languages are not selected for training. Oketunji et al. \cite{oketunji2023large} argue that LLMs inherently generate bias, and propose a large language model bias index to quantify and address biases, enhancing the reliability of LLMs.  Li et al. \cite{li2023fairness} introduced a systematic methodology to evaluate fairness and bias that could be displayed in LLMs, where a number of biased prompts are fed into LLMs and the probabilistic metrics for both individuals and groups that indicate the level of fairness are computed. In the domain of education, reinforced statements like "You should be unbiased for the sensitive feature (race or gender in experiments)" are helpful in mitigating the biased responses from LLMs. Chhikara et al. \cite{chhikara2024fewshot} show some gender bias from LLMs and explore possible solutions using few-shot learning as well as retrieval augmented generation. Caliskan et al. \cite{caliskanpsychological} examine social bias among scholars by evaluating LLMs (llama 2, Yi, Tulu, etc.) with various input prompts and argue that fine-tuning is the most effective approach to maintaining fairness. Li et al. \cite{li2024large} think that LLMs frequently present dominant viewpoints while ignoring alternative perspectives from minority parties (underrepresented in the training data), resulting in potential biases. They proposed a FAIRTHINKING pipeline to automatically generate roles that enable LLMs to articulate diverse perspectives for fair expressions. Li et al. \cite{li2024steering} analyzes reasoning bias in decision-making systems of education and health care and devises a guided-debiasing framework that incorporates a prompt selection mechanism. 

\vspace{-3pt}
\subsection{Reliability and Safety}\vspace{-1mm}
LLMs have encountered reliability issues, including hallucinations, the production of toxic output, and inconsistencies in responses. These challenges are particularly significant in the educational sector. Hallucinations, where LLMs generate fictitious content, are a critical concern highlighted by Ji et al. \cite{ji2023survey}. Zhuo et al. \cite{zhuo2023red} have outlined ethical considerations regarding the potential of LLMs to create content containing offensive language and explicit material. Cheng et al. \cite{cheng2024dated} have discussed the issue of temporal misalignment in LLM data versions, introducing a novel tracer to track knowledge cutoffs. Shoaib et al. \cite{shoaib2023deepfakes} underscore the risks of misinformation and disinformation through seemingly authentic content, suggesting the adoption of cyber-wellness education to boost public awareness and resilience. Liu et al. \cite{liu2024sora} explore the application of text-to-video models, like Sora, as tools to simulate real-world scenarios. They caution, however, that these models, despite their advanced capabilities, can sometimes lead to confusion or mislead students due to their limitations in accurately representing physical realism and complex spatial-temporal contexts. To improve the reliability of LLMs, Tan et al. \cite{tan2024tuningfree} have developed a metacognitive strategy that enables LLMs to identify and correct errors autonomously. This method aims to detect inaccuracies with minimal human intervention and signals when adjustments are necessary. Additionally, the use of retrieval augmented generation (RAG) has been identified by Gao et al. and Zhao et al. as an effective way to address the issues of hallucinations and response inconsistencies, improving the reliability and accuracy of LLMs in content generation \cite{gao2024retrievalaugmented, zhao2024retrievalaugmented}.

\vspace{-5pt}
\subsection{Transparency and Accountability}\vspace{-1mm}
LLM, by design, runs as a black-box mechanism, so it comes with transparency and accountability concerns. Milano et al. \cite{Milano2023} and BaHammam et al. \cite{bahammam2023adapting} raise several challenges about LLMs to higher education, including plagiarism, inaccurate reporting, cheating in exams, as well as several other operational, financial, pedagogical issues. As a further thought of students using generative AI in homework or exams, Macneil et al. \cite{macneil2024imagining} discuss the respective impact to the traditional assessment methods, and think we educators should come up with new assessment framework to take into account the usage of Chat-GPT-like tools. Zhou et al. \cite{zhou2024the} brought up academic integrity ethical concerns that specifically confused teachers and students and calls for rethinking of policy-making. As specific measures, Gao et al. \cite{gao2024llmasacoauthor} introduce a novel concept called mixcase, representing a hybrid text form involving both machine-generated and human-generated and developed detectors that can distinguish human and machine texts. To tackle LLM ethical concerns in intellectual property violation, Huang et al. \cite{huang2023citation} propose incorporating citation while training LLMs, which could help enhance content transparency and verifiability. Finlayson, et al. \cite{finlayson2024logits} developed a systematic framework to efficiently discover the LLM's hidden size, obtaining full-vocabulary outputs, detecting and disambiguating different model updates, which could help users hold providers accountable by tracking model changes, thus enhancing the accountability.

\vspace{-3pt}
\subsection{Privacy and Security}\vspace{-1mm}
Privacy and security protection have become increasingly important topics with the rise of LLMs, especially in the education sector where they deserve heightened scrutiny. Latham et al. \cite{Latham2019} conducted a case study to explore the general public's perceptions of AI in education, revealing that while research has largely focused on the effectiveness of AI, critical areas such as learner awareness and acceptance of tracking and profiling algorithms remain underexplored. This underscores the need for more research into the ethical and legal aspects of AI in education. Das et al. \cite{das2024security} conducted an extensive review on the challenges of protecting personally identifiable information in the context of LLM use, highlighting widespread security and privacy concerns. Shoaib et al. \cite{shoaib2023deepfakes} addressed the threats to personal privacy posed by deepfake content, proposing solutions like the use of detection algorithms and the implementation of standard protocols to bolster protection. Ke et al. \cite{ke2024exploring}raised concerns about data privacy and the ethical implications of employing LLMs in psychological research, emphasizing the importance of safeguarding participant privacy in research projects. This highlights the necessity for researchers to understand the limitations of LLMs, adhere to ethical standards, and consider the potential consequences of their use. Suraworachet et al. \cite{Suraworachet_2024} provided a comparative analysis of student information disclosure using LLMs versus traditional methods. Their findings point to challenges in valid evaluation, respecting privacy, and the absence of meaningful interactions in using LLMs to assess student performance. In terms of mitigation strategies, Hicke et al. \cite{hicke2023aita} suggested frameworks that combine Retrieval-Augmented Generation (RAG) and fine-tuning techniques for enhanced privacy protection. Meanwhile, Masikisiki et al. \cite{masikisiki2023investigating} highlighted the significance of offering users the option to delete their interactions, emphasizing the importance of user control over personal data.

\vspace{-3pt}
\subsection{Overly Dependency on LLMs}\vspace{-1mm}
Given the impressive performance of LLM generative ability, there is great concern that students blindly rely on LLMs for most of their work, leaving their ability to think independently disappearing. Milano et al. \cite{Milano2023} discuss the problem of overreliance caused by Chat-GPT-like applications that students may use to compose essays and academic publications without improving their writing skills, which is essential to cultivate critical thinking. The concern might have even more impact on foreign-language students or students who are educationally disadvantaged while placing less emphasis on learning how to craft well-written texts. 
Krupp et al. \cite{krupp2023challenges} discuss the challenges of overreliance on the implications of LLMs in education and propose some moderated approaches to mitigate such effects. Similarly, Zuber et al. \cite{zuber2023vox} discuss the risk that overreliance can bring to democracy and suggest cultivating thinking skills in children, fostering coherent thought formulation, and distinguishing between machine-generated output and genuine. They argue that LLMs should be used to augment but not substitute human thinking capacities. Adewumi et al. \cite{adewumi2023procot} also present scenarios that students tend to rely on LLMs for essay writing rather than writing their own, and demonstrates that using a probing-chain-of-thought tool could substantially stimulate critical thinking in the accompany with LLMs.

  \vspace{-5pt}
\section{Future Directions}
\begin{figure}
    \centering
    \includegraphics[width=0.4\textwidth]{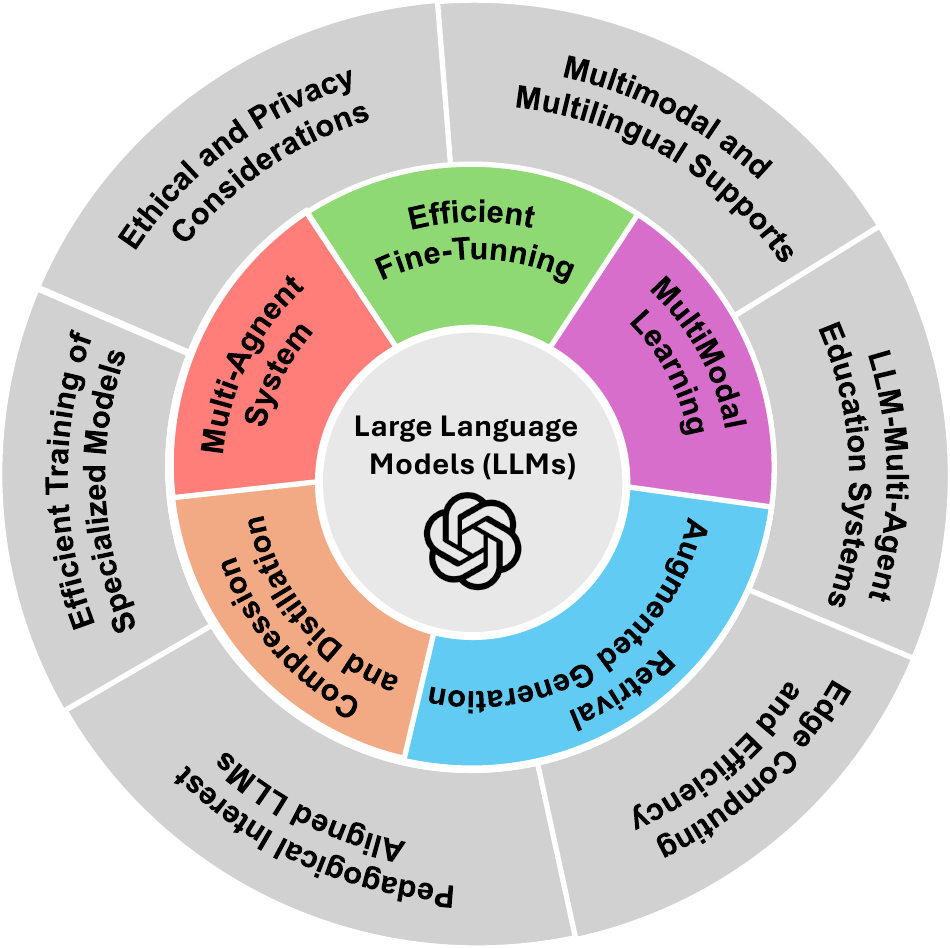}\vspace{-1mm}
    \caption{Future directions for LLMs in education.}\vspace{-2mm}
    \label{fig:future}
\end{figure}

Here, we discuss the future opportunities for LLMs in education and summarize the promising directions in Figure~\ref{fig:future}. For each direction, we discuss the potential application of advanced LLM-based techniques and conclude their impacts on the future of education.

\subsection{Pedagogical Interest Aligned LLMs}

Although advanced LLMs like GPT-4 have demonstrated promising performance while being applied to experimental applications in education, applying LLMs directly in real-world instruction is still challenging, as delivering high-quality education is a complicated task that involves multi-discipline knowledge and administration constraints \cite{milano2023large}. To solve these issues, future researchers in this direction could leverage advanced generation techniques like Retrieval-Augmented Generation (RAG) \cite{gao2023retrieval} to inform LLMs with the necessary prior information and guide LLMs to generate the pedagogical interest-aligned results. Apart from that, collecting large-size pedagogical instruction datasets from real-world instruction scenarios and fine-tuning existing LLMs to align with human instructors' behavior will also be an interesting direction to explore for future research. With learning from the preference of human instructors, LLMs could encoder the pedagogical restriction and knowledge patterns within LLMs' parameter space, and generate pedagogical interest-aligned results without too much intervention from the external information.

\subsection{LLM-Multi-Agent Education Systems}

The generous usage of LLMs in language comprehensive, reasoning, planning, and programming inspires works like AutoGen \cite{wu2023autogen} in developing a collaboration framework that involves multiple LLMs to solve complicated tasks through conversation-format procedures. Similarly, problems in education commonly involve multi-step processing logic, which will be a good fit for using the multi-agent-based LLMs system. The recent work by \citet{yang2024content} has demonstrated the great potential of the multi-agent framework for grading tasks. In this work, a human-like grading procedure is achieved by leveraging multiple LLM-based grader agents and critic agents, and the discrepancy errors made by individual judges are corrected through the group discussion procedure. For future research in this direction, more types of LLM-based agents can be included and their functions may spread from specific command executors to high-level plan makers. More importantly, human instructors, who are viewed as the special agents in the system, can also get involved with LLMs' interaction directly and provide any necessary interventions to the system flexibly.


\subsection{Multimodal and Multilingual Supports}

The high similarity between different human languages naturally enables LLMs to effectively support multilingual tasks. In addition to that, the recent findings of the alignment between multimodal and language tokens \cite{wu2023next} further extend LLMs beyond textual analysis, venturing into multimodal learning analytics. By accepting diverse inputs, LLMs can take advantage of the mutual information between different data resources and provide advanced support to challenging tasks in education. For future works in multimodal direction, more attention could be placed on developing LLMs capable of interpreting and integrating these varied data sources, offering more nuanced insights into student engagement, comprehension, and learning styles. Such advances could pave the way for highly personalized and adaptive learning experiences that are tailored to the unique needs of each student. On the other hand, multilingual LLMs provide convenient access to quality global education resources for every individual with his/her proficient language. With developing robust models that not only translate, but also understand cultural nuances, colloquial expressions, and regional educational standards, researches in this direction would help learners around the world to benefit from LLMs in their native languages and significantly improve equity and inclusion in global education.

\subsection{Edge Computing and Efficiency}\vspace{-1mm}
Integrating LLMs with edge computing presents a promising avenue to enhance the efficiency and accessibility of educational technologies. By processing data closer to the end user, edge computing can reduce latency, increase content delivery speed, and enable offline access to educational resources. Future efforts could explore optimizing LLMs for edge deployment, focusing on lightweight models that maintain high performance while minimizing computational resources, which would be particularly beneficial in areas with limited internet connectivity, ensuring equitable access to educational tools. Additionally, processing data locally reduces the need to transmit sensitive information over the Internet, enhancing privacy and security. Edge computing could be a potential framework to utilize LLMs while adhering to stringent data protection standards.
\vspace{-5pt}
\subsection{Efficient Training of Specialized Models}\vspace{-1mm}
The development of specialized LLMs tailored to specific educational domains or subjects represents a significant opportunity for future research. This direction involves creating models that not only grasp general language understanding, but also possess deep knowledge in fields such as mathematics, science, or literature. The point is that specialized LLMs could achieve a deep understanding of specific subjects, offering insights and support that are highly relevant and accurate but also more cost-effective. The challenge lies in the efficient training of these models, which requires innovations in data collection, model architecture, and training methodologies. Specialized models could offer more accurate and contextually relevant help, improving the educational experience for both students and educators.
\vspace{-5pt}
\subsection{Ethical and Privacy Considerations}\vspace{-1mm}
Ethical and privacy considerations take center stage as LLMs become increasingly integrated into educational settings. Future research must address the responsible use of LLMs, including issues related to data security, student privacy, and bias mitigation. The development of frameworks and guidelines for ethical LLM deployment in education is crucial. This includes ensuring transparency in model training processes, safeguarding sensitive information, and creating inclusive models that reflect the diversity of the student population. Addressing these considerations is essential to build trust and ensure the responsible use of LLMs in education.

\vspace{-5pt}
\section{Conclusion}
The rapid development of LLMs has revolutionized education. In this survey, we provide a comprehensive review of LLMs specifically applied for various educational scenarios from a multifaceted taxonomy, including student and teacher assistance, adaptive learning, and miscellaneous tools. In addition, we also summarize the related datasets and benchmarks, as well as current challenges and future directions. We hope that our survey can facilitate and inspire more innovative work within LLMs for education.

\bibliographystyle{ACM-Reference-Format}
\bibliography{06-reference}

\appendix
\section{Appendix}

Table \ref{tab:dataset} summarizes the commonly used publicly available datasets and benchmarks for evaluating LLMs on education applications.

\begin{table*}
    \centering
    \caption{Summary of existing datasets and benchmarks in the area of LLMs for education.}\vspace{-3.5mm}
    \label{tab:dataset}
    \resizebox{0.9\textwidth}{!}{
    \begin{tabular}{ccccccccc}
    \hline
      Dataset\&Benchmark  & App  & User & Subject & Level & Language & Modality & 
      Amount & Source\\
    \hline
         GSM8K        &QS   & student & math          & K-12              & EN     & text & 8.5K   & \cite{cobbe2021training}\\
         MATH         &QS   & student & math          & K-12              & EN     & text & 12.5K  & \cite{hendrycks2021measuring}\\
         Dolphin18K   &QS   & student & math          & K-12              & EN     & text & 18K    & \cite{huang2016well}\\
         DRAW-1K      &QS   & student & math          & comprehensive           & EN     & text & 1K     & \cite{upadhyay2016annotating}\\
         Math23K      &QS   & student & math          & K-12              & ZH     & text & 23K    & \cite{wang2017deep}\\
         Ape210K      &QS   & student & math          & K-12              &EN, ZH  & text & 210K   & \cite{zhao2020ape210k}\\
         MathQA       &QS   & student & math          & K-12              & EN     & text & 37K    & \cite{amini2019mathqa}\\
         ASDiv        &QS   & student & math          & K-12              & EN     & text \& image & 2K    & \cite{miao2021diverse}\\
         IconQA       &QS   & student & math          & K-12              & EN     & text \& table & 107K  & \cite{lu2021iconqa}\\
         TQA          &QS   & student & science       & K-12              & EN     & text \& image & 26K   & \cite{kim2018textbook}\\ 
         Geometry3K   &QS   & student & geometry      & K-12              & EN     & text \& image & 3K    & \cite{lu2021inter}\\
         AI2D         &QS   & student & science       & K-12              & EN     & text \& image & 5K    & \cite{kembhavi2016diagram}\\
         SCIENCEQA    &QS   & student & science       & K-12              & EN     & text \& image & 21K   & \cite{chen2023theoremqa} \\ 
         MedQA        &QS   & student & medicine      & professional      & EN     & text          & 40K   & \cite{jin2021disease} \\ 
         MedMCQA      &QS   & student & medicine      & professional      & EN     & text          & 200K  & \cite{pal2022medmcqa} \\ 
         TheoremQA    &QS   & student & science       & college           & EN     & text          & 800   & \cite{chen2023theoremqa} \\
         Math-StackExchange &QS & student &math       & comprehensive     & EN     & text          & 310K  & \cite{yuan2020automatic} \\
         TABMWP       &QS   & student & math          & K-12              & EN     & text          & 38K   & \cite{lu2022dynamic}\\
         ARC          &QS   & student & comprehensive & comprehensive     & EN     & text          & 7.7K  & \cite{clark2018think} \\
         C-Eva        &QS   & student & comprehensive & comprehensive     & ZH     & text          & 13.9K & \cite{huang2024c} \\
         GAOKAO-bench &QS   & student & comprehensive & comprehensive     & ZH     & text          & 2.8K  & \cite{Zhang2023EvaluatingTP}\\
         AGIEval      &QS   & student & comprehensive & comprehensive & EN, ZH & text          & 8k    & \cite{zhong2023agieval}\\
         MMLU         &QS   & student & comprehensive & comprehensive     & EN     & text          & 1.8K  & \cite{hendrycks2020measuring} \\
         CMMLU        &QS   & student & comprehensive & comprehensive     & ZH     & text          & 11.9K & \cite{li2023cmmlu} \\
         SuperCLUE    &QS   & student & comprehensive & comprehensive     & ZH     & text          & 15.9K & \cite{xu2023superclue} \\
     \hline
         LANG-8       &EC   &student  & linguistic    & language training & Multi   & text         & 1M    & \cite{rothe2021simple}\\ 
         CLANG-8      &EC   &student  & linguistic    & language training & Multi   & text         & 2.6M   &\cite{rothe2021simple}\\ 
         CoNLL-2014   &EC   &student  & linguistic    & language training & EN      & text         & 58k & \cite{ng2014conll}\\ 
         BEA-2019     &EC   &student  & linguistic    & language training & EN      & text         & 686K &\cite{bryant2019bea}\\ 
         SIGHAN       &EC   &student  & linguistic    & language training & ZH      & text         & 12K &\cite{tseng2015introduction}\\ 
         CTC          &EC   &student  & linguistic    & language training & ZH      & text         & 218K &\cite{zhao2022overview}\\ 
         FCGEC        &EC   &student  & linguistic    & language training & ZH      & text & 41K & \cite{xu2022fcgec}\\ 
         FlaCGEC      &EC   &student  & linguistic    & language training & ZH      & text &13K &\cite{du2023flacgec}\\ 
         GECCC        &EC   &student  & linguistic    & language training & CS      & text &83K &\cite{naplava2022czech} \\ 
         RULEC-GEC    &EC   &student  & linguistic    & language training & RU      & text &12K & \cite{rozovskaya2019grammar} \\ 
         Falko-MERLIN &EC   &student  & linguistic    & language training & GE      & text &24K & \cite{grundkiewicz2014wiked}\\ 
         COWS-L2H     &EC   &student  & linguistic    & language training & ES      & text &12K &\cite{davidson2020developing} \\ 
         UA-GEC       &EC   &student  & linguistic    & language training & UK      & text &20K &\cite{syvokon2021ua}\\ 
         RONACC       &EC   &student  & linguistic    & language training & RO      & text &10K &\cite{cotet2020neural}\\ 
         
         Defects4J    &EC   &student  & computer science  & professional & EN \& Java          & text\& code & 357    & \cite{just2014defects4j}\\
         ManyBugs     &EC   &student  & computer science  & professional & EN \& C             & text\& code &185    & \cite{le2015manybugs}\\
         IntroClass   &EC   &student  & computer science  & professional & EN \& C             & text\& code & 998    & \cite{le2015manybugs}\\
         QuixBugs     &EC   &student  & computer science  & professional & EN \& multi& text\& code &40     & \cite{lin2017quixbugs}\\
         Bugs2Fix     &EC   &student  & computer science  & professional & EN \& Java    & text\& code      & 2.3M   & \cite{tufano2019empirical}\\
         CodeReview   &EC   &student  & computer science  & professional & EN \& Multi    &text\& code     & 642    & \cite{li2022automating}\\
         CodeReview-New&EC  &student  & computer science  & professional & EN \& Multi    &text\& code     & 15     & \cite{guo2024exploring}\\
      
      \hline   
         SciQ         &QG   & teacher & science       & MOOC              & EN     & text          & 13.7K & \cite{welbl2017crowdsourcing} \\ 
         RACE         &QG   & teacher & linguistic    & K-12              & EN     & text          &  100K & \cite{lai2017race}\\ 
         FairytaleQA  &QG   & teacher & literature    & K-12              & EN     & text          &  10K  & \cite{xu2022fantastic} \\ 
         LearningQ    &QG   & teacher & comprehensive & MOOC              & EN     & text          &  231K & \cite{chen2018learningq}\\ 
         KHANQ        &QG   & teacher & science       & MOOC              & EN     & text          &  1K   & \cite{gong2022khanq} \\ 
         EduQG        &QG   & teacher & comprehensive & MOOC              & EN     & text          &  3K   & \cite{hadifar2023eduqg}  \\ 
         MCQL         &QG   & teacher & comprehensive & MOOC              & EN     & text          &  7.1K & \cite{liang2018distractor} \\ 
         Televic      &QG   & teacher & comprehensive & MOOC              & EN     & text          &  62K  & \cite{bitew2022learning} \\ 
       \hline  
         CLC-FCE      &AG   & teacher & linguistic    & standardized test & EN     & text          &  1K   & \cite{yannakoudakis2011new}\\
         ASAP         &AG   & teacher & linguistic    & K-12              & EN     & text          &  17K  & \cite{tigina2023analyzing} \\
         TOEFL11      &AG   & teacher & linguistic    & standardized test & EN     & text          &  1K    & \cite{blanchard2013toefl11} \\
         ICLE         &AG   & teacher & linguistic    & standardized test & EN     & text          &  4K    & \cite{stab2014annotating} \\
         HSK          &AG   & teacher & linguistic    & standardized test & ZH     & text          &  10K   & \cite{cui2011principles} \\
    \hline
    \end{tabular}
    } \vspace{-3mm}
\end{table*}


\end{document}